\documentclass[12pt]{article}
\usepackage{concmath}
\usepackage[OT1]{fontenc}
\usepackage[total={6in, 9in}]{geometry}
\usepackage{graphicx}
\usepackage{amsmath}
\usepackage{amssymb} 
\usepackage{booktabs} 
\usepackage[round]{natbib}
\usepackage{authblk}
\usepackage{placeins}
\usepackage{setspace}
\usepackage{xcolor}

\title{Pre-screening breast cancer with machine learning and deep learning}

\author[*]{Rolando Gonzales Martinez}
\author[**]{Daan-Max van Dongen}
\affil[*]{Royal Netherlands Academy of Arts and Sciences, Netherlands Interdisciplinary Demographic Institute (NIDI), University of Groningen (Netherlands). Contact: \textcolor{blue}{r.m.gonzales.martinez@rug.nl}}
\affil[**]{University College London (United Kingdom)} 

\date{January 2023}

\begin{document}

\maketitle
\begin{abstract}
\noindent Deep learning has been widely applied in breast cancer screening to analyze images obtained from X-rays, ultrasounds, magnetic resonances, and biopsies. We suggest that deep learning can also be used for pre-screening cancer by analyzing demographic and anthropometric information of patients, as well as biological markers obtained from routine blood samples and relative risks obtained from meta-analysis and international databases. In this study, we applied feature selection algorithms to a database of 116 women, including 52 healthy women and 64 women diagnosed with breast cancer, to identify the best pre-screening predictors of cancer. We utilized the best predictors to perform $k$-fold Monte Carlo cross-validation experiments that compare deep learning against traditional machine learning algorithms. Our results indicate that a deep learning model with an input-layer architecture that is fine-tuned using feature selection can effectively distinguish between patients with and without cancer, since the average area under the curve (AUC) of a deep learning model is 87\%, with a 95\% confidence interval between 82\% and 91\%. Additionally, compared to machine learning, deep learning has the lowest uncertainty in its predictions, as indicated by its standard deviation of AUC equal to 0.0345. These findings suggest that deep learning algorithms applied to cancer pre-screening offer a radiation-free, non-invasive, and affordable complement to screening methods based on imagery. The implementation of deep learning algorithms in cancer pre-screening offer opportunities to identify individuals who may require imaging-based screening, can encourage self-examination, and decrease the psychological externalities associated with false positives in cancer screening. The integration of deep learning algorithms for both screening and pre-screening will ultimately lead to earlier detection of malignancy, reducing the healthcare and societal burden associated to cancer treatment.
\end{abstract}

\newpage
\doublespacing

\section{Introduction}

Breast cancer is now the most commonly diagnosed cancer worlwide. More than 2.26 million new cases of breast cancer were estimated for 2020 \citep{wilkinson2022understanding}, with Belgium and the Netherlands having the highest age-standardized incidence of breast cancer, and developing countries as Somalia and Syria having the highest breast cancer mortality. Machine learning and deep learning algorithms are at the core of breast cancer screening in developing countries \citep{torres2019comparison}. These algorithms are applied to predict the presence of anomalies related to the presence breast cancer in digitalized images \citep{debelee2020survey,yu2021deep, zhou2020lymph} obtained from magnetic resonance imaging, ultrasounds \citep{zheng2020deep}, digital breast tomosynthesis \citep{bai2021applying}, breast density from mammograms \citep{yala2019deep}, tissue images \citep{naik2020deep}, or cell nuclei from fine needle aspirates of breast mass \citep{khuriwal2018breast, street1993nuclear}. 

We suggest that deep learning can also be applied to the prognosis of breast cancer through the application of these algorithms to the information of patients obtained from medical records (demographic and anthropometric information), biological markers obtained from routine blood samples, and relative risks obtained from meta-analysis and publicly available databases. The development and progression of breast cancer has been linked for example to glucose dysmetabolism, insulin resistance and changes in adipokine secretion \citep{crisostomo2016hyperresistinemia}, and hence predictive models of breast cancer based on biological markers (glucose, insulin, HOMA, leptin, adiponectin, resistin and MCP-1) can be applied as pre-screening tools for the early detection of breast cancer, particularly when the information of biological markers is combined with the demographic and anthropometric information of patients and their relative risk by age and sex according to body mass index (BMI). While the carcinogenic mechanisms that related overweight and obesity to breast cancer are still uncertain, the positive association between high BMI and breast cancer risk in postmenopausal women was speculated to result from the higher level of estrogen derived from the aromatization of androstenedione within the larger fat reserves of women with high BMI \citep{lake1997women}. The tetrad of BMI, leptin, the ratio of leptin/adiponectin and the antigen 15.3 (CA15-3), are considered reliable biomarkers of breast cancer \citep{santillan2013tetrad}, and resistin levels tend to be significantly elevated in postmenopausal breast cancer, after adjusting for demographic, metabolic and clinicopathological features \citep{dalamaga2013serum}. 

In this study, we analyzed the ability of machine learning and deep learning to predict the presence of breast cancer with pre-screening information. Machine learning and deep learning algorithms were applied to the information of 116 women, 52 healthy women and 64 women that were diagnosed with breast cancer at the University Hospital Centre of Coimbra (UHCC). The UHCC data contains demographic and anthropometric information, besides biological markers obtained from blood samples of the 166 women that were part of the study. The UHCC dataset was extended with information of relative risks of breast cancer obtained from the Global Burden of Disease Study 2019 \citep{vos2020global} and the relative risks of the meta-analysis of \cite{liu2018association}. The optimal predictors of breast cancer were selected with the SULOV-gradient boosting algorithm. The SULOV algorithm reduced the full set of potential explanatory predictors to a subset of features with the highest relevance and the lowest redundancy for predicting breast cancer. The optimal set of predictors were used in a $k$-fold Monte Carlo cross-validation experiment that compared deep learning against seven traditional machine learning algorithms: support vector machines, neural networks, logistic regression, XGBoost, random forests, naive Bayes and stochastic gradient algorithms. The results indicate that deep learning has the highest predictive ability and the highest precision, compared to machine learning algorithms. This finding indicates that deep learning algorithms can be used to create a non-invasive, radiation-free and affordable pre-screening medical recommendation system that can complement and guide cancer screening. 

\section{Materials and Methods}

\subsection{Data}

The data of the University Hospital Centre of Coimbra (UHCC) contains information of 116 women, 52 healthy women and 64 women that were diagnosed with breast cancer. The information of UHCC was collected between 2009 and 2013 for a study of biomarkers of breast cancer, based on the results of routine blood analysis. Demographic and anthropometric information of patients was recorded during the 
first consultation of the patients with the physician \citep{patricio2018using}. The demographic information in the dataset is the age of women and the anthropometric information is the BMI of women. The diagnosis of breast cancer was the result of a mammography and was confirmed histologically with samples collected before surgery  or treatment. The samples of tumor tissue were obtained by mastectomy or tumourectomy and were evaluated by a pathologist at the Anatomic Pathology Department of UHCC. The counterfactual group of women without breast cancer in the database were healthy volunteers that were enrolled in the study as controls. Both the patients with breast cancer and the control group of healthy women had no prior cancer treatment and were free from infections, acute diseases or comorbidities at the time of participating in the study. 

Blood samples of the women with breast cancer and the control group of healthy women were extracted at the Laboratory of Physiology of the Faculty of Medicine of University of Coimbra from peripheral venous blood vials. The blood samples were collected after an overnight fasting. The fasting blood was centrifuged (2500 g) at 4$^{\circ}$C and stored at -80$^{\circ}$C for biochemical determinations of serum glucose levels (mg/dL), insulin levels ($\mu$U/mL), and serum values of leptin (ng/mL), adiponectin ($\mu$g/mL), resistin (ng/mL), and chemokine monocyte chemoattractant Protein 1 (MCP-1, (pg/dL)). An Homeostasis Model Assessment (HOMA) index that measures insulin resistance was calculated with the information of the fasting insulin level ($\mu$U/mL) and fasting glucose level (mmol/L). 

\begin{figure}[ht]
\centering
\includegraphics[width=15cm]{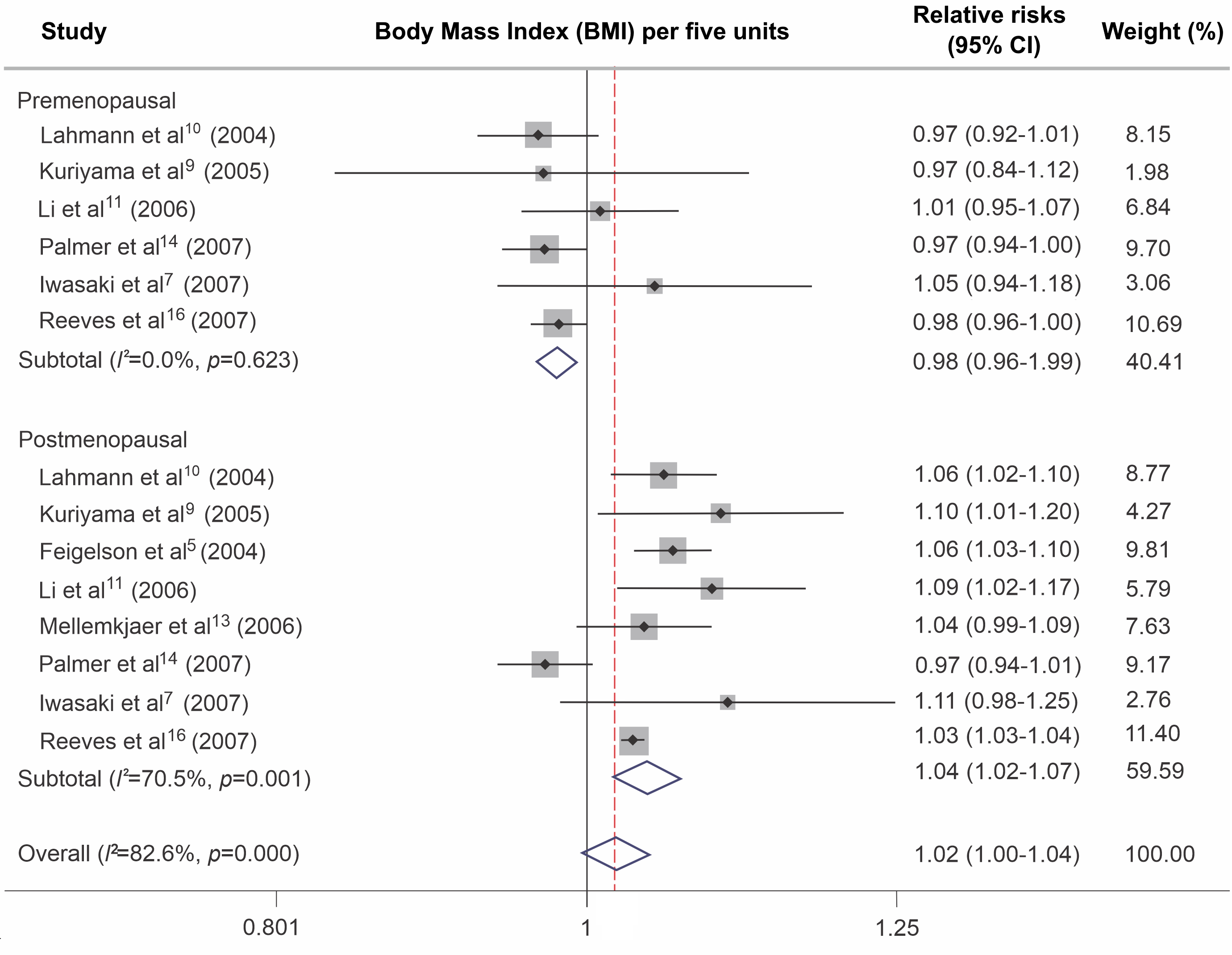}
\caption{Subgroup meta-analysis of the association between BMI increment (per five units) and breast cancer risk, by menopausal status. Reproduced from  \cite{liu2018association}. \label{fig:meta}}
\end{figure}

The information of the UHCC was extended with data of relative risks of breast cancer by age obtained from the Global Burden of Disease (GBD) Study 2019 \citep{vos2020global} and the relative risks of breast cancer associated to body mass index (BMI) from the dose-response meta-analysis of of \cite{liu2018association}. The GBD Study 2019 was coordinated by the Institute for Health Metrics and Evaluation, who estimated the burden of diseases, injuries, and risk factors for 204 countries, territories and selected subnational locations worldwide. The GBD database contains information of relative risks of cancer for high BMI (BMI $\geq$ 25), by age and sex. The dose-response meta-analysis of \cite{liu2018association} was based on 12 prospective cohort studies comprising 22,728,674 participants. The results of \cite{liu2018association} show that every 5 kg/m2 increase in BMI corresponds to a 2\% increase in breast cancer risk in women, with differential results for premenopausal women, for which higher BMI could be a protective factor in breast cancer risk (Figure \ref{fig:meta}). Table \ref{tab:desc} shows the descriptive statistics of the complete dataset used in this study; this is, the original UHCC data that was extended with relative
risks (RRs) from publicy available databases and studies. 

\begin{table}[ht] 
\centering
\caption{Descriptive statistics of the database\label{tab:desc}}
\begin{tabular}{lcccc}
\toprule
\textbf{}	& \textbf{mean}	& \textbf{std.dev.} & \textbf{min} & \textbf{max} \\
\midrule
Age (years) &	57.30 &	16.11 &	24 &	89 \\
BMI &	27.58 &	5.02 &	18.37 &	38.58 \\
Glucose (mg/dL) &	97.79 &	22.53 &	60.00 &	201.00 \\
Insulin ($\mu$U/mL) &	10.01 &	10.07 &	2.43 &	58.46 \\
HOMA &	2.69 &	3.64 &	0.47 &	25.05 \\
Leptin (ng/mL) &	26.62 &	19.18 &	4.31 &	90.28 \\
Adiponectin ($\mu$g/mL) &	10.18 &	6.84 &	1.66 &	38.04 \\
Resistin (ng/mL) &	14.73 &	12.39 &	3.21 &	82.10 \\
RRs (Liu et al., 2018) & 1.51 &	0.48 &	0.40 &	2.44 \\
RRs GBD (center)    & 1.01 &	0.10 &	0.89 &	1.09 \\
RRs GBD (lower)     & 0.97 &	0.08 &	0.87 &	1.04 \\
RRs GBD (upper)     & 1.05 &	0.11 &	0.91 &	1.14 \\
High BMI (binary)   & 0.34 &	0.47 &	0 &	1 \\
Obesity (binary)    & 0.32 &	0.47 &	0 &	1 \\
\bottomrule
\end{tabular}
\end{table}

\subsection{Methods}

Experiments using $k$-fold Monte Carlo cross-validation were conducted to evaluate the ability that machine learning and deep learning classifiers have to differentiate between patients with and without cancer, based on demographic and anthropometric information, as well as biomarkers and relative risks in relation to age and body mass index. Feature selection with the SULOV algorithm was applied to determine the optimal set of predictors for identifying breast cancer.

Deep learning is a machine learning approach based on multiple layers of artificial neural networks that are densely-connected. Each layer transforms the input data from the previous layer into a new representation trough non-linear functions connected by synaptic weights \citep{shen2017deep}. This allows the deep learning model to learn both locally and in the inter-relationships of the whole data, through a hierarchical structure \citep{haque2020deep}. As deep learning can handle imbalanced, heterogeneous, and high dimensional data, and is robust to changes in the input data due its multiple hidden layers, it is at the center of artificial intelligence and is  increasingly being used to mine large and complex relationships hidden in biomedical data \citep{dash2022deep, baldi2018deep}. Previous applications of deep learning in biomedical science include image reconstruction and the automated interpretation of downstream images in biomedical optics \citep{tian2021deep}, segmentation of medical images in diseased and healthy regions \citep{haque2020deep,isensee2021nnu}, photoacoustic imaging of chromophores \citep{grohl2021deep}, gene expression \citep{he2020integrating}, protein structure prediction \citep{torrisi2020deep}, brain machine interfaces \citep{bozhkov2020deep}, and clinical language processing \citep{ionescu2020deep}. The deep learning architectures applied in this study have rectified networks and L1 and L2 regularization in the kernel, bias and activation functions of the dense hidden layers.

A recursive minimum redundancy maximum relevance (MRMR) algorithm of SULOV-gradient boosting was applied for feature selection \citep{ramirez2017fast}. Feature selection is the process of reducing the number of input variables when developing a predictive model \citep{brownlee2020data}. Feature selection is a preprocessing step that improves model performance by identifying the variables that are the most relevant to the prediction task, consequentially removing irrelevant and redundant variables from the set of potential predictors. Irrelevant features are those that can be removed without affecting learning performance, while redundant features are correlated features that are individually relevant, but due to their co-presence the removal of one of them does not affect learning performance \citep{liu2007computational}. MRMR finds the minimal-optimal subset of features through the selection of highly predictive but uncorrelated features. The algorithm starts calculating the mutual information score of all the pairs of highly correlated variables and then selects the ones with the highest Information scores and least correlation with each other. This first step is called the \textsc{SULOV} (Searching for Uncorrelated List Of Variables) algorithm. In the second step, XGBoost is used to repeatedly find the best features to predict the target variable in a test sample using the model estimated in the train sample with the variables selected with SULOV in step 1. XGBoost is a regularized gradient boosting algorithm  based on boosted tree ensembles \citep{chen2016xgboost} that has proven to be highly successful solving a vast array of predictive problems due to its ability to handle the bias-variance trade-off \citep{nielsen2016tree}.  In biomedical science, the MRMR algorithm has been used before for the feature selection of temporal gene expression data \citep{ding2005minimum, radovic2017minimum}.

The performance of the machine learning and deep learning classifiers was evaluated using the Area Under the Receiver Operating Characteristic Curve (AUC). The performance of the deep learning algorithm---measured by AUC---was compared against the AUC of seven machine learning algorithms: XGBoost, stochastic gradient, support vector machines, random forests, neural networks, naive Bayes and logistic regression. AUC measures the ability of the model to distinguish between positive and negative classes; in this case, the presence of breast cancer (positive class) or the absence of breast cancer (negative class). AUC ranges between 0 and 1, with a value of 1 indicating a perfect classifier and a value of 0.5 indicating a classifier no better than random guessing. AUC is particularly useful when the distribution of classes in the target variable is imbalanced, because in the presence of unbalanced class distribution, a model could achieve high accuracy by simply predicting the majority class most of the time. AUC is a better metric in these cases because it takes into account both the true positive rate and the false positive rate of the predictive model. The AUC of the machine learning and deep learning algorithms was evaluated through Monte Carlo experiments based on a $k$-fold cross-validation that splits the data in $k$-train and $k$-test samples. The machine learning and deep learning classifiers are estimated with the data of the train sample, and the predictive ability of the models is tested in the test sample that was not used to estimate the algorithms. 

\section{Results}

\subsection{Feature selection with SULOV-gradient boosting}

Table~\ref{tab:sulov} shows the results of selecting the best predictors of breast cancer with the SULOV algorithm. The recursive MRMR-SULOV algorithm was implemented
for values of the correlation threshold $\rho$ of features for values of $\rho$ between $\rho = 0.01$ and $\rho = 0.99$, in order to mitigate the impact  of different correlation values on the selection of optimal predictors and produce results that are robust to the choice of the correlation threshold between variables. The variables that were selected more frequently by the SULOV algorithm are age, resistin, the upper values of the relative risks of breast cancer published by the GBD database, glucose, adiponectin, high BMI, MCP-1,leptin, the relative risks of breast cancer of \cite{liu2018association}, obesity and insulin levels in the blood samples (Table~\ref{tab:sulov}). These are the less correlated variables and at the same time the most relevant features selected by the MRMR SULOV-gradient boosting. A previous independent study based on fuzzy neural networks \citep{silva2019using} and stochastic vector machines \citep{patricio2018using} also found resistin, glucose and age relevant for breast cancer prediction. Age is considered a particularly  relevant risk factor for breast cancer since the diagnosis of this neoplastic disease is most frequently found in women in menopausal transition and less frequently found in women below 45 years of age \citep{kaminska2015breast}. Even for premenopausal women, previous studies \citep{kresovich2019methylation} have found that each 5-year acceleration in biological age corresponds with a 15\% increase in breast cancer risk. 

In relation to the biomarkers selected by the SULOV algorithm---resistin, adiponectin, and glucose---, both the low serum adiponectin levels and high resistin levels were found to be associated with increased breast cancer risk previously \citep{kang2007relationship}. Resistin has been significantly associated with tumor and inflammatory markers, cancer stage, tumor size, grade and lymph node invasion \citep{dalamaga2013serum}, since resistin facilitates breast cancer progression via TLR4-mediated induction of mesenchymal phenotypes and stemness properties \citep{wang2018resistin}. Moreover, circulating resistin associated with the presence of breast cancer in a dose–response manner appears to have adiposity-independent roles in breast carcinogenesis \citep{sun2010adipocytokine}. There is also evidence that glucose and other factors related to glucose metabolism, such as insulin and insulin-like growth-factors, may contribute to breast cancer development, in pre and post menopausal women \citep{sieri2012prospective}, because malignant cells extensively use glucose for proliferation \citep{muti2002fasting}. In contrast, the homeostasis model assessment index (HOMA) is frequently excluded from the set of features by the SULOV algorithm due to its low relevance and high correlation with other features. This result is expected and suggests that the SULOV-gradient boosting algorithm properly excludes redundant features, because HOMA is a measure of insulin resistance measurement that is calculated as a combination of fasting insuline and glucose levels and was found also to be correlated with BMI in previous studies \citep{timoteo2014optimal}.

\begin{table}[ht] 
\centering
\caption{Recursive MRMR feature selection with SULOV-gradient boosting\label{tab:sulov}}
\begin{tabular}{lcccc}
\toprule
\textbf{}	& \textbf{Frequency}	& \textbf{min $\rho$} & \textbf{mean $\rho$} & \textbf{std.dev. $\rho$} \\
\midrule
Age (years)         &	97 &	0.00 &	0.49 &	0.28 \\
Resistin (ng/mL)    &	90 &	0.00 &	0.51 &	0.28 \\
RRs GBD (upper)     &	77 &	0.05 &	0.55 &	0.27 \\
Glucose (mg/dL)     &	76 &	0.21 &	0.59 &	0.22 \\
Adiponectin ($\mu$g/mL) & 75 &	0.23 &	0.60 &	0.22 \\
High BMI (binary)   &	65 &	0.27 &	0.64 &	0.20 \\
MCP-1 (pg/dL)       &	65 &	0.26 &	0.65 &	0.19 \\
Leptin (ng/mL)      &	64 &	0.30 &	0.65 &	0.19 \\
RRs (Liu et al., 2018) & 59 &	0.37 &	0.68 &	0.17 \\
Obesity (binary)    &	59 &	0.11 &	0.55 &	0.28 \\
Insulin ($\mu$U/mL) &	33 &	0.05 &	0.53 &	0.26 \\
BMI                 &	30 &	0.62 &	0.81 &	0.11 \\
HOMA                &	29 &	0.69 &	0.83 &	0.09 \\
RRs GBD (center)    &	 8 &	0.86 &	0.90 &	0.04 \\
RRs GBD (lower)     &	 8 &	0.21 &	0.81 &	0.25 \\
\bottomrule
\end{tabular}
\end{table}

 \subsection{Machine learning and deep learning results}

The best predictors selected by the SULOV algorithm were used in the deep learning and machine learning algorithms. In the deep learning algorithm, a sigmoid activation function was used for the output layer of the neural networks, due to the binary nature of the target variable (presence or not of breast cancer). The optimization of the deep learning model was performed with adaptive moment estimation \citep{kingma2014adam} during $3 \times 10^2$ epochs, with a batch size equal to $1 \times 10$, using a binary cross-entropy loss function. Grid search was applied to choose the type of activation function, the number of hidden layers, and the number of nodes in the hidden layers of the neural networks. The highest predictive power was obtained with a rectified linear activation function (ReLU) in the input layer, four hidden layers with $1 \times 10^2$ nodes, a L1-L2 regularizer in the kernel function and a L2 regularizer for the bias and the activation functions of the hidden units. The better performance of ReLU activation functions compared to tangent-hyperbolic or (S)eLU activations functions \citep{clevert2015fast, klambauer2017self} was also expected, since the use of piece-wise linear hidden units based on the ReLU activation functions is considered a major algorithmic change that improved the performance of feed-forward networks \citep{goodfellow2017deep} and reduced the computational burden of calculating the exponential function in activation functions \citep{glorot2011deep}. 

\begin{figure}[ht]
\centering
\includegraphics[width=12 cm]{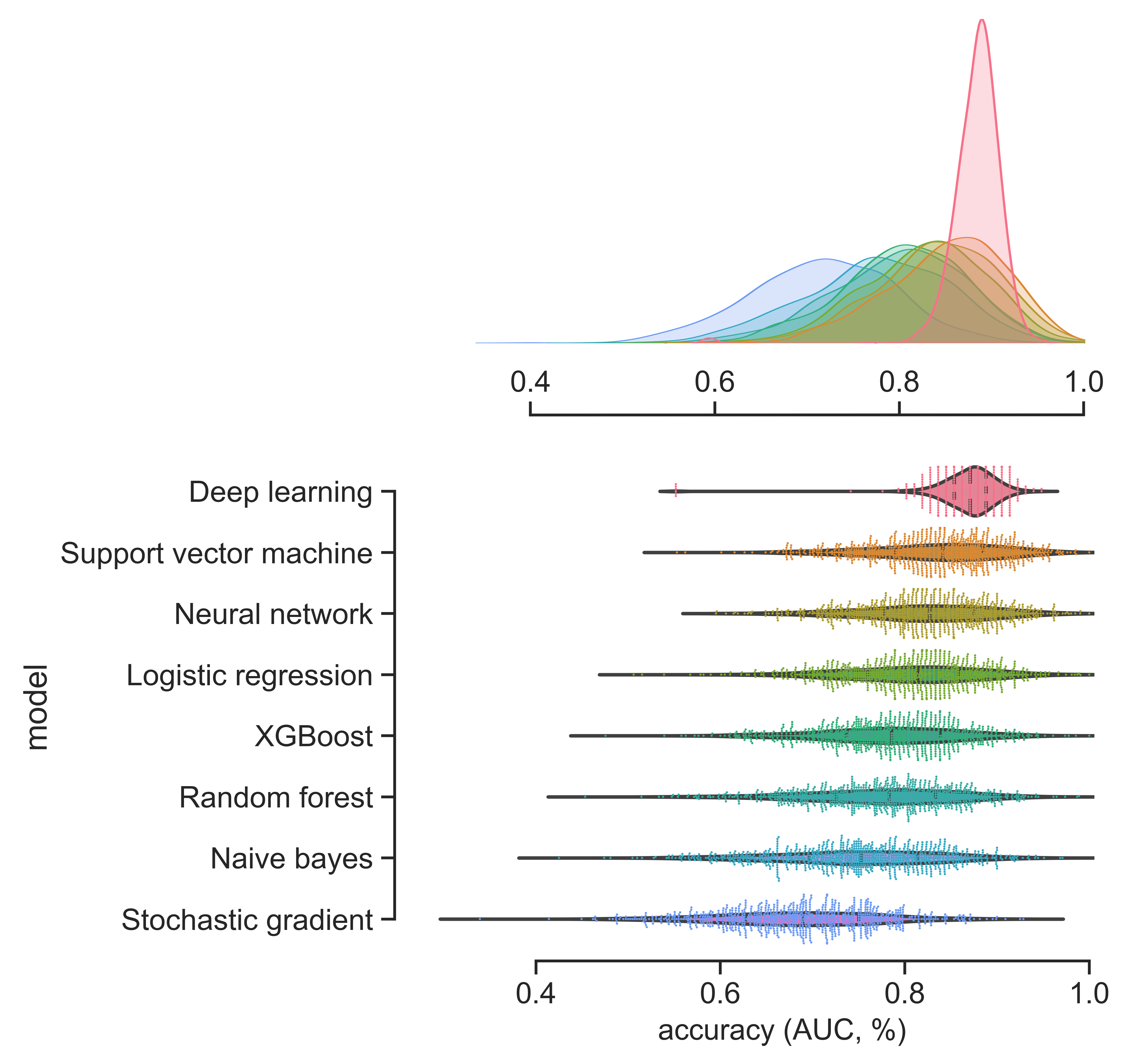}
\caption{AUC of machine learning and deep learning classifiers in the Monte Carlo cross-validation\label{fig:MLDL}}
\end{figure}

Table \ref{tab:resMLDL} and Figure \ref{fig:MLDL} show the results of predicting the presence of breast cancer with machine learning and deep learning algorithms. On average, the highest predictive ability to differentiate between women with and without breast cancer is obtained with deep learning. 
The deep learning algorithm has on average an AUC equal to 87\%, with a 95\% confidence interval between 82\% and 91\% (Table \ref{tab:resMLDL}). Support vector machines, in turn, have a lower average AUC (83\%) with a wider confidence interval (95\% CI = [68\%, 95\%]), and neural networks with a simple architecture have an average AUC of 82\%, also with a wider confidence interval (95\% CI = [67\%, 95\%]). Compared to the traditional machine learning algorithms, the deep learning algorithms also have the lowest predictive uncertainty, measured by the standard deviation and the percentiles of the distribution of the AUC obtained with the $k$-fold Monte Carlo experiments (Table \ref{tab:resMLDL}). The lowest dispersion of the AUC (Figure \ref{fig:MLDL}) is obtained with the deep learning algorithm (AUC standard deviation = 0.0345), followed by support vector machines (AUC standard deviation = 0.0711) and neural networks (AUC standard deviation = 0.0720). This last results shows the high precision of predictions obtained with deep learning compared to machine learning.

\FloatBarrier

\begin{table}[ht] 
\centering
\caption{AUC of the machine learning \newline and deep learning algorithms\label{tab:resMLDL}}
\begin{tabular}{lcccc}
\toprule
\textbf{Model}	& \textbf{mean}	& \textbf{std.dev.} & \textbf{p2.5} & \textbf{p97.5} \\
\midrule
Deep learning			& 0.8699 & 0.0345 & 0.8190 & 0.9138 \\
Support vector machines	& 0.8344 & 0.0711 & 0.6768 & 0.9527 \\
Neural network			& 0.8232 & 0.0720 & 0.6667 & 0.9476 \\
Logistic regression		& 0.8078 & 0.0741 & 0.6569 & 0.9334 \\
XGBoost					& 0.7834 & 0.0755 & 0.6263 & 0.9191 \\
Random forest			& 0.7763 & 0.0804 & 0.6035 & 0.9192 \\
Naive bayes				& 0.7504 & 0.0869 & 0.5686 & 0.9001 \\
Stochastic gradient		& 0.6861 & 0.0860 & 0.5120 & 0.8553 \\
\bottomrule
\end{tabular}
\end{table}
\unskip

\FloatBarrier

\section{Discussion}

Breast cancer is a prevalent form of cancer among women, with early detection being crucial for survival. The screening of breast cancer is typically performed through methods such as mammograms, ultrasound, magnetic resonance imaging (MRI), self-examination, or examination by a clinician. Advances in deep learning have led to the development of algorithms that can detect anomalies related to breast cancer in images obtained from patients, particularly for the segmentation and classification of normal and abnormal breast tissue from thermograms \citep{mohamed2022deep, torres2019comparison}. 

Our study explored the application of deep learning for pre-screening breast cancer. We applied machine learning and deep learning to observational data that contains demographic and anthropometric information of cancer patients and a control group of healthy women, in addition to biological markers obtained from routine blood samples and relative risks of cancer obtained from publicly available databases published by international studies. We found that deep learning algorithms have the highest predictive ability to distinguish between women with and without breast cancer. Deep learning also has the lowest uncertainty---the highest precision---in its predictions compared to machine learning. 

Our findings indicate that pre-screening cancer with deep learning algorithms offers a non-invasive, radiation-free, and affordable alternative for the early detection of breast cancer. A medical recommendation system of pre-screening based on deep learning can aid in the early detection of mass anomalies that may not yet be detectable through self-examination or have not yet caused symptoms, but are at a stage where they are easier to treat. This is particularly relevant for breast cancer, as previous studies have shown that women tend to have an already developed metastatic breast cancer when they are first diagnosed, and they have a poor prognosis regardless of their menopausal status \citep{kang2007relationship}. Additionally, pre-screening breast cancer could reduce the risks associated with traditional screening methods, such as false positive test results and unnecessary tests that are expensive, invasive, time-consuming, and may cause anxiety in patients \citep{lerman1991psychological, mathioudakis2019systematic}.

The integration of deep learning algorithms for cancer pre-screening in medical recommendation systems can guide the frequency and follow-up of radiographic imaging and tissue testing, particularly in developing countries, where access to traditional cancer screening and medical personnel is limited for low-income populations in rural areas, making breast cancer a leading cause of death due to late detection \citep{kakileti2020personalized}. Future studies should investigate the potential of increasing the predictive ability of deep-learning algorithms by incorporating additional cancer risk factors, such as those associated with smoking and alcohol consumption, since a recent analysis of the impact of 34 risk factors for 23 cancer types suggests that smoking, alcohol use, and high BMI are the leading contributors for 4.45 million cancer deaths (44.4\% of all cancer deaths) globally in 2019 \citep{tran2022global}.
\bigskip

\bibliography{biblio.bib}

\end{document}